\documentclass[]{interact}
\pdfoutput=1
\usepackage{epstopdf}% To incorporate .eps illustrations using PDFLaTeX, etc.
\usepackage[caption=false]{subfig}% Support for small, `sub' figures and tables
\usepackage[numbers]{natbib}
\usepackage{hyperref}					% Reference
\hypersetup{colorlinks,%				% Reference color setup	
	linkcolor=blue,%
	citecolor=blue}

\usepackage{algorithm,algpseudocode}
\usepackage{multirow}
\usepackage{array}
\usepackage{graphicx}
\usepackage{color}

\theoremstyle{plain}% Theorem-like structures provided by amsthm.sty

\theoremstyle{definition}

\theoremstyle{remark}

\begin{document}
\captionsetup[figure]{labelfont={bf},labelformat={default},labelsep=period,name={Fig.}}
%\articletype{ARTICLE TEMPLATE}% Specify the article type or omit as appropriate

\title{Remote sensing image fusion based on Bayesian GAN}

\author{
\name{Junfu Chen\textsuperscript{a}, Yue Pan\textsuperscript{a} and Yang Chen\textsuperscript{a}}
\affil{\textsuperscript{a}College of computer science and technology; Nan jing University of Aeronautics \& Astronautics; Nan jing; 211106; China }
}

\maketitle

\begin{abstract}

Remote sensing image fusion technology (pan-sharpening) is an important means to improve the information capacity of remote sensing images. Inspired by the efficient arameter space posteriori sampling of Bayesian neural networks, in this paper we propose a Bayesian Generative Adversarial Network based on Preconditioned Stochastic Gradient Langevin Dynamics (PGSLD-BGAN) to improve pan-sharpening tasks. Unlike many traditional   generative models that consider only one optimal solution (might be locally optimal), the proposed PGSLD-BGAN performs Bayesian inference on the network parameters, and explore the generator posteriori distribution, which assists selecting the appropriate generator parameters. First, we build a two-stream generator network with PAN and MS images as input, which consists of three parts: feature extraction, feature fusion and image reconstruction. Then, we leverage Markov discriminator to enhance the ability of generator to reconstruct the fusion image, so that the result image can retain more details. Finally, introducing Preconditioned Stochastic Gradient Langevin Dynamics policy, we perform Bayesian inference on the generator network. Experiments on QuickBird and WorldView datasets show that the model proposed in this paper can effectively fuse PAN and MS images, and be competitive with even superior to state of the arts in terms of subjective and objective metrics.
\end{abstract}


\section{Introduction}

With the rapid development of earth observation technology, earth remote sensing has entered the era of multi-platform, multi-sensor, and multi-angle observation. The demand of acquiring remote sensing data with high spatial resolution and high spectral resolution has further boomed. However, because the optical remote sensing system is limited by the optical diffraction limit, modulation transfer function, signal-to-noise ratio and so on, it is quite arduous to obtain satellite remote sensing images with both high spatial and high spectral properties. In order to alleviate this problem, many optical earth observation satellites carry two kinds of optical sensors to obtain panchromatic (PAN) images and multispectral (MS) images, through which relevant experts can fuse the different but complementary characteristics. The Pan-sharpening method is a significant branch of remote sensing image fusion. Its essence is to fuse multispectral images (MS) with low spatial information and high spectral quality and panchromatic images (PAN) with high spatial quality and low spectral information ,so as to achieve multi-spectral images with both high spatial resolution and high spectral resolution.

So far, researchers have proposed many methods for fusion of panchromatic and multispectral images. Traditional panchromatic and multispectral algorithms can be roughly classified into two categories: component substitution (CS) methods and am\'elioration de la r\'esolution spatiale par injection de structures(ARSIS) concept methods. When the component substitution method is used for image fusion, the multi-spectral image is projected into a new feature space. Then, the intensity component is obtained and the panchromatic image is used to replace the intensity component, which improves the spatial resolution of the multi-spectral image. Component substitution method generally consists of up-sampling, forward transformation, intensity matching, component substitution and inverse transformation \citep{p01}. Generally, the component substitution method is easy to implement with high algorithm efficiency and the fusion result can reach a higher spatial resolution. However, for the low-frequency information of the image keeping changing during the fusion process, it will inevitably cause spectral distortion \citep{p02}. Representative algorithms of CS methods include PCA (principal component analysis) method \citep{p03,p04} and IHS (intensity, hue, saturation)methods \citep{p05,p06}.

ARSIS can overcome the problem of spectral distortion, but it causes some spatial degradation. ARSIS method mainly extracts the spatial information of the panchromatic image through muti-scale methods or filters, and injects the extracted spatial information into the original muti-spectral image to improve the spatial resolution of the multi-spectral image. Comparing with the component substitution method, the performance of the muti-resolution analysis method is easily affected by the image registration result, and is prone to aliasing effects and edge artifacts. Representative algorithms of ARSIS methods include wavelet transform-based methods \citep{p07}, Laplacian Pyramidal decomposition methods \cite{p08}, and filtering-based methods \citep{p09,p10}. 

In recent years, the application of deep learning, especially the depth model based on convolutional neural network, to process panchromatic and multi-spectral image fusion has become a research hotspot.Masi et al.\citep{p11} proposed a pan-sharpening method based on the three-layer CNN architecture originally designed for image super resolution \citep{p12}.Zhong et al. \citep{p13} proposed a novel two-stage RSIF algorithm. In the first stage, the algorithm also uses similar methods \citep{p12} to improve the spatial resolution of LMS images. Liu et al. \citep{p14} proposed a dual-stream fusion network (TFNet) to solve the problem of generic sharpening.

With the successful application of traditional deep learning models in fields such as image segmentation, classification and fusion, many scholars are gradually introducing the concept of Bayesian learning into deep learning. On the other hand, Bayesian deep learning can be used to explore the uncertainty of the model and assist the model to make decisions.
Inspired by this, we propose a Bayesian Generative Adversarial Network based on Preconditioned Stochastic Gradient Langevin Dynamics (PGSLD-BGAN) to improve the pan-sharpening performance of remote sensing images. Similar to most pan-sharpening using an encoder structure, the generative network of PGSLD-BGAN serves as a fusion network that encodes the features of PAN and MS and then decodes the fused features to implement image fusion. The image results generated by the generative network are input to the discriminator network as fake data, while the corresponding reference images (high-resolutin MS images) are co-trained as real data.

To sum up, our contribution can be concluded as the following three aspects:
1. We propose a Generative Network Adversarial structure to perform the pan-sharpening task. The proposed model can conduct the image reconstruction better by discriminating the network as compared to the common encoder structure.

2. We propose a PSGLD approach for Bayesian learning on GAN, which leverages a RMSPROP optimizer and train the model more stably than SGHMC methods to avoid the vanishing gradient problem.

3. To our knowledge, our work is the first to employ the Bayesian deep learning methodology to select better generator parameters for pan-sharpening by sampling the model parameter posterior. The proposed model has an excellent description of latent variable space of the model than traditional deep generative models.

\section{Related works}
This section briefly introduces some related works, including remote sensing image fusion technology based on deep learning and the development of Bayesian neural network.

\subsection{Deep learning methods for image fusion}
In recent years, deep learning has been successfully applied to remote sensing image fusion due to its strong ability to extract image features. Masi et al. \citep{p11} first tried to apply the SRCNN framework \citep{p12} to the pan-sharpening problem. In order to adapt to the input of the SRCNN framework, Masi et al. superimposed the up-sampled MS and PAN to form a five-band image. Zhong et al. \citep{p13} also used SRCNN for remote sensing image fusion. The difference is that they divide pan-sharping into two steps. The first step uses SRCNN to enhance the resolution of MS, and the second step applies GS transformation to complete pan-sharping.
In order to avoid network degradation caused by the increase in the number of network layers, researchers have also explored the use of residual network to improve the performance of remote sensing image fusion. For example, Rao et al. \citep{p15} proposed a residual CNN model to conduct pan-sharpening. Comparing with the traditional CNN model, this method achieves fast convergence and high pan-sharpening quality. Yang et al. \citep{p16} proposed a deep network structure called PanNet, which can incorporate knowledge of a specific field. In particular, this approach can learn in the high-pass filter domain instead of the image domain to preserve spatial information. Liu et al. \citep{p14} proposed a two-stream fusion network (TFNet) to solve the pan-sharpening problem. This method employs two networks to extract features from PAN images and MS images respectively, and then integrates them together. Finally, the desired high spatial resolution MS image is achieved from the fused features through the image reconstruction network.

\subsection{Bayesian neural networks}
From the perspective of probability theory, the existing neural network can be interpreted as point estimation, and the use of point estimation to determine the network weight of the classification learning task cannot measure uncertainty. Hinton et al.\citep{p17} are the first to propose combining the scalability, flexibility and predictive performance of neural networks with Bayesian uncertainty modeling. Neal \citep{p18} showed that under certain assumptions, as the width of the shallow BNN increases, its limit distribution can be regarded as a Gaussian process. In recent years, as the popularity of deep learning has risen, a variety of mathematical methods have emerged to implement Bayesian deep learning. For example, Chen et al. \citep{p19} proposed a Hamiltonian Monte Carlo method based on stochastic gradient to implement Bayesian sampling algorithm. In order to improve the natural implementation of stochastic approximation, they introduced the second-order Langevin dynamics. Blundell \citep{p20} introduced a new network propagation method Bayes By Backprop to realize Bayesian neural network. Gal \citep{p21} et al. proposed a new theoretical framework, using drop-out in deep neural networks as a Bayesian inference method, providing a simple and easy to implement uncertainty modeling technique. Maddox et al. \citep{p22} proposed a simple and scalable approach called SWA-Gaussian based on stochastic weight averaging to explore the uncertain representation and calibration in Bayesian deep learning.

\section{PSGLD-BGAN}
Before introducing PSGLD-BGAN, we briefly review the related content of Generative Adversarial Network (GAN) and a general framework of Bayesian GAN.

\subsection{Generative Adversarial Network}
GAN \citep{p23} is a generative model, which can generate real data by implicitly modeling the distribution of high-dimensional data. Inspired by the idea of game theory, GAN trains a pair of competing networks at the same time through a mini-max game process. To further illustrate, A generator G and a discriminator D compete with each other to achieve Nash equilibrium. The goal of the generator is to capture the potential distribution in the actual data and generate artificial data samples. The purpose of the discriminator D is to distinguish the generated fake samples and real data samples. In the training process, the quality of the generated samples and the discriminative ability of the discriminator have been interactively improved.
GAN conducts sampling from the real data distribution $p_{data}\left( x \right) $, and samples $z$ from the prior distribution $p_z\left( z \right) $. The input of $G$ is a random vector $z$, the generated data is recorded as $G\left( z \right)$, and its distribution is $p_g\left( z \right) $. The purpose of GAN is to make the distribution of $p_g\left( z \right) $ and the real data sample $p_{data}\left( x \right)$ similar. The input of $D$ can be real data $x$ or generated data $G\left( z \right)$. The output result of $D$ is scaled to [0,1]. When the input is real data $x$, the output value $D\left( x \right)$ will approach 1, and when the input is generated data $G\left( z \right)$, the output probability value $D\left( G\left( z \right) \right) $ will approach 0. Eq. (\ref{e1}) describes the objective function of GAN.
\begin{equation}
\label{e1}
\underset{G}{\min}\underset{D}{\max}L\left( G,D \right) =E_{x\sim p_{data}\left( x \right)}\left[ \log D\left( x \right) \right] +E_{z\sim p_z\left( x \right)}\left[ \log \left( 1-D\left( G\left( z \right) \right) \right) \right] 
\end{equation}
\subsection{Bayesian GAN}
Saatci et al.(23) introduced the concept of Bayesian GAN for the first time, and utilized Stochastic Gradient HMC (SGLD) to perform posterior sampling to obtain the marginal distribution of GAN parameters. The given hyperparameters $\alpha _d$ and $\alpha _g$ are used to control the discriminator parameter $\theta _d$ and the generator parameter $\theta _g$ respectively. In other words, we set $p\left(\theta _d|\alpha _d \right)$ and $p\left(\theta _g|\alpha _g\right)$ as the prior probability of the discriminator parameter and the generator parameter. Bayesian GAN specially designs the likelihood to maximize the possibility of conditional distribution and equivalently optimize the corresponding goal of GAN . Instead of the traditional GAN based on the maximum likelihood estimation of point quality, it marginalizes the distribution of the generator (multimodal itself) to better explore the distribution of multimodal data. Eq. (\ref{e2}) and Eq. (\ref{e3}) show the joint distribution modeling process of discriminator and generator.
\begin{equation}
\label{e2}
p\left( \theta _g|\theta _d \right) \propto \exp \left\{ -L_g\left( \theta _g;\theta _d \right) \right\} p\left( \theta _g|\alpha _g \right) 
\end{equation}
\begin{equation}
\label{e3}
p\left( \theta _d|\theta _g \right) \propto \exp \left\{ -L_g\left( \theta _d;\theta _g \right) \right\} p\left( \theta _d|\alpha _d \right)
\end{equation}
Literature \citep{p24} also designed a general framework for Bayesian Gan to update parameters using Monte Carlo sampling. The specific steps are shown in Eq. (\ref{e4}).
\begin{equation}
\label{e4}
\left\{ \theta _{g,k}^{\left( t+1 \right)} \right\} _{k=1}^{K}\sim p\left( \theta _d|\left\{ \theta _{d,k}^{\left( t \right)} \right\} _{k=1}^{K} \right) =\left( \prod_k{p\left( \theta _g|\theta _{d,k}^{\left( t \right)} \right)} \right) ^{\frac{1}{K}}=\exp \left\{ -\frac{1}{K}\sum_k{L_g\left( \theta _g;\theta _{d,k} \right)} \right\} p\left( \theta _g|\alpha _g \right)
\end{equation}
\begin{equation}
\label{e5}
\left\{ \theta _{d,k}^{\left( t+1 \right)} \right\} _{k=1}^{K}\sim p\left( \theta _g|\left\{ \theta _{d,k}^{\left( t \right)} \right\} _{k=1}^{K} \right) =\left( \prod_k{p\left( \theta _g|\theta _{g,k}^{\left( t \right)} \right)} \right) ^{\frac{1}{K}}=\exp \left\{ -\frac{1}{K}\sum_k{L_d\left( \theta _d;\theta _{g,k} \right)} \right\} p\left( \theta _d|\alpha _d \right)
\end{equation}
\subsection{Bayesian inference with PSGLD}
Unlike the SGHMC approach adopted in comparison to the literature \citep{p24}, we propose a Bayesian GAN based on Preconditioned Stochastic Gradient Langevin Dynamics\citep{p25}. Stochastic Gradient Langevin Dynamics\citep{p26} naturally incorporates Stochastic gradient algorithms and Langevin dynamics. What's more, SGLD enables us to capture uncertainty in a Bayesian learning manner while applying it efficiently to large datasets. This approach implements Bayesian learning on neural networks by adding appropriate noise to random gradients in an annealing stepwise manner.
\begin{equation}
\label{e6}
\Delta \theta _t=\frac{\epsilon _t}{2}\left( \nabla \log p\left( \theta _t \right) +\frac{N}{n}\sum_{i=1}^n{\nabla \log p\left( x_{ti}|\theta _t \right)} \right) +\eta _t
\end{equation}
where $\frac{\epsilon _t}{2}$ represents a sequence of steps and satisfies the properties $\sum_{t=1}^{\infty}{\epsilon_{t}=\infty}$ , $\sum_{t=1}^{\infty}{\epsilon _{t}^{2}}=\infty$, and $\eta_{t}<\infty$. $\left(0,\epsilon _{t} \right)$, $\mathbf{X}^t=\left\{ \boldsymbol{x}_{t1},...,\boldsymbol{x}_{tn} \right\} $ denotes the set of subsets divided from the full dataset, corresponding to the batches during the training of the neural network.

In order to accommodate rapid changes in curvature, Li et al. \citep{p25} used a user-chosen preconditioning matrix $G_\theta$ based on the SGLD.Considering the family of probability distributions $p\left( X|\theta\right)$ from a Riemannian manifold perspective, they take the non-Euclidean geometry through this perspective to direct the sampling method. Then, Eq. (\ref{e6}) can then be rewritten as \label{e7}.

\begin{equation}
\label{e7}
\Delta \theta _t=\frac{\epsilon _t}{2}\left[ G\left( \theta _t \right) \left( \nabla \log p\left( \theta _t \right) +\frac{N}{n}\sum_{i=1}^n{\nabla \log p\left( x_{ti}|\theta _t \right)} \right) +\Gamma \left( \theta _t \right) \right] +G^{\frac{1}{2}}\left( \theta _t \right) \eta _t
\end{equation}
where $\Gamma \left( \mathbf{\theta }_t \right) =\sum_j{\frac{\partial G_j\left( \boldsymbol{\theta } \right)}{\partial \theta _j}}$ describes the change pattern of the preconditioning matrix $G\left( \theta\right)$.
The preconditioner strategy in equation d is implemented by the stochastic optimizer RMSprop. this preconditioner is sequentially updated by the current gradient information, which is expressed as follows.
\begin{equation}
\label{e8}
G\left( \theta _{t+1} \right) =\text{diag} \left( 1 \oslash \left( \lambda +\sqrt{V\left( \theta _{t+1} \right)} \right) \right)
\end{equation}
\begin{equation}
\label{e9}
V\left( \theta _{t+1} \right) =\alpha V\left( \theta _t \right) +\left( 1-\alpha \right) \bar{g}\left( \theta _t;\mathbf{X} \right) \odot \bar{g}\left( \theta _t;\mathbf{X} \right) 
\end{equation}
where $\bar{g}\left( \boldsymbol{\theta }_t;\mathbf{X}^t \right)$ represents the mean of the gradient of a mini-batch $D^t$, which can be used for simplification into $\frac{1}{n} \sum\limits_{i=1}^n{\nabla _{\theta}\log p\left( \boldsymbol{x}_{ti}|\boldsymbol{ \theta }_t \right)}$. The symbols $\odot$ and $\oslash$ represent elment-wise matrix multiplication and division, respectively.

Therefore, we can give a general PSGLD-BGAN training framework, denoted as Algorithm 1.
Algorithm 1 Preconditioned SGLD Bayesian GAN.
\begin{algorithm}
\caption{$ \textbf{:}$ SGLD Bayesian GAN}
\label{alg1}
\hspace*{0.02in}{\bf Input:}
$\left\{ \epsilon _t \right\} _{t=1:T},\lambda ,\alpha$\\
\hspace*{0.02in}{\bf Output:}
$\left\{ \theta _{d,t} \right\} _{t=1:T},\left\{\theta _{g,t} \right\} _{t=1:T}$\\
\hspace*{0.02in}{\bf Initialize:}
$\mathbf{V}_0\gets 0,random\,\,\theta _{g,1},\theta _{d,1}$\\
\begin{algorithmic}[1]
\For{$\,\text{epoch\,\,}t\gets 1:T$}
	\State $\text{Sample\,\,a\,\,}\min\text{ibatch\,\,of\,\,size\,\,n,\,\,}\mathbf{X}^t=\left\{ \boldsymbol{x}_{t1},...,\boldsymbol{x}_{tn} \right\} $.
	\State $\text{Sample\,\,a\,\,}\min\text{ibatch\,\,of\,\,}n\,\,\text{noise\,\,samples}\left\{ \boldsymbol{z}^{\left( 1 \right)},...,\boldsymbol{z}^{\left( n \right)} \right\} \,\,from\,\,noise\,\,prior\,\,p\left( z \right)$.
	\State $\text{Update\,\,}p\left( \theta _g|\theta _d \right) \,\,\text{through\,\,preconditioned\,\,SGLD\,\,RMSprop:}$
	\State $Estimate\,\,gradient\,\,\bar{g}\left( \theta _{g,t};X^t \right) =\frac{1}{n}\sum\limits_n^1{\nabla \log p\left( x_{ti};\theta _{gt}|z^{\left( i \right)},\theta _{dt} \right)}$
	\State $V\left( \theta _{g,t} \right) \gets \alpha V\left( \theta _{g,t-1} \right) +\left( 1-\alpha \right) \bar{g}\left( \theta _{g,t};\mathbf{X}^t \right) \odot \bar{g}\left( \theta _{g,t};\mathbf{X}^t \right) $
	\State $G\left( \theta _{g,t} \right) \gets \text{diag}\left( 1\oslash\left( \lambda +\sqrt{V\left( \theta _t \right)} \right) \right) $
	\State $ \theta _{g,t+1}\gets \theta _{g,t}+\frac{\epsilon _t}{2}\left[ G\left( \theta _{g,t} \right) \left( \nabla \log p\left( \theta _t \right) +N\bar{g}\left( \theta _{g,t} \right) \right) +\Gamma \left( \theta _{g,t} \right) \right] +G^{\frac{1}{2}}\left( \theta _{g,t} \right) \eta _t$
	\State $\text{Sample\,\,a\,\,}\min\text{ibatch\,\,of\,\,}n\,\,\text{noise\,\,samples}\left\{ \boldsymbol{z}^{\left( 1 \right)},...,\boldsymbol{z}^{\left( n \right)} \right\} \,\,from\,\,noise\,\,prior\,\,p\left( z \right)$.
	\State $\text{Update\,\,}p\left( \theta _d|\theta _g \right) \,\,\text{through\,\,preconditioned\,\,SGLD\,\,RMSprop:}$
	\State $Estimate\,\,gradient\,\,\bar{g}\left( \theta _{d,t};X^t \right) =\frac{1}{n}\sum\limits_n^1{\nabla \log p\left( x_{ti};\theta _{dt}|z^{\left( i \right)},\theta _{gt} \right)}$
	\State $V\left( \theta _{d,t} \right) \gets \alpha V\left( \theta _{d,t-1} \right) +\left( 1-\alpha \right) \bar{g}\left( \theta _{d,t};\mathbf{X}^t \right) \odot \bar{g}\left( \theta _{d,t};\mathbf{X}^t \right) $
	\State $G\left( \theta _{d,t} \right) \gets \text{diag}\left( 1\oslash\left( \lambda +\sqrt{V\left( \theta _{d,t} \right)} \right) \right)$
	\State $\Delta \theta _{d,t+1}\gets \theta _{d,t}+\frac{\epsilon _t}{2}\left[ G\left( \theta _{d,t} \right) \left( \nabla \log p\left( \theta _t \right) +N\bar{g}\left( \theta _{d,t} \right) \right) +\Gamma \left( \theta _{d,t} \right) \right] +G^{\frac{1}{2}}\left( \theta _{d,t} \right) \eta _t$
\EndFor
\end{algorithmic}
\end{algorithm}

\section{Image fusion task through proposed PSGLD-BGAN}
In this section, we will describe how to perform remote sensing image fusion tasks using proposed PSGLD-BGAN. The network design and the loss function are described in detail. 
\subsection{Motivation and purpose}
For multi-source remote sensing images, PAN images contain high-resolution spatial information, and MS images contain the multispectral features of the image. At present, deep learning models for remote sensing image fusion has become the mainstream method. Unlike traditional methods, which tend to cause problems such as spectral distortion and information loss, deep learning models can perform feature extraction while retaining more feature information, so high-quality fusion images can be obtained. However, traditional deep learning methods use point estimation methods to form a fusion image generator, which lacks exploration of the potential performance capabilities of the generator. For this reason, we employ PGSLD to perform Bayesian learning on GAN and apply PGSLD-BGAN to the field of pan-sharping.

\begin{figure}  
	\centering  
	\includegraphics[width=10cm]  {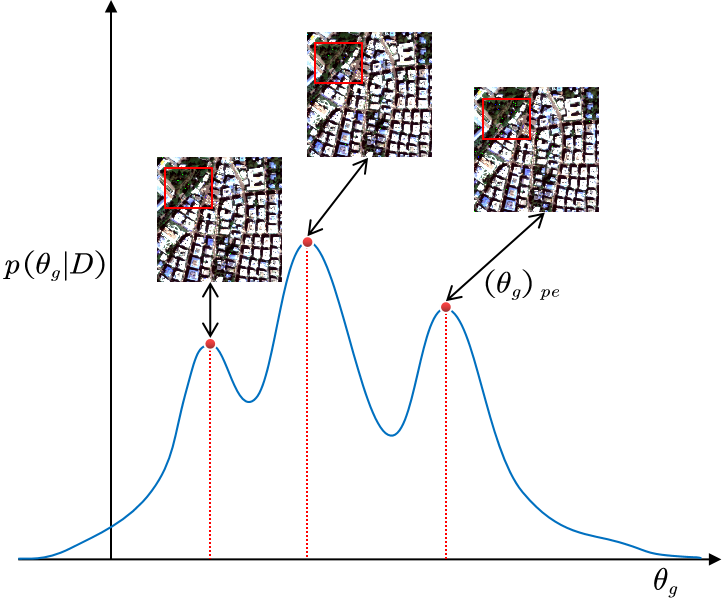} 
	\caption{\label{1} Representation of posterior over the generator parameter space}
	\label{Fig1}
\end{figure}

Figure. \ref{Fig1} describes the concept of the posterior distribution of Bayesian learning on pan-sharping. Each $\theta_g$ on the abscissa corresponds to the generator parameter hypothesis. We show there examples generated by three parameter settings, where $\theta_g$ corresponds to the image fusion result output by traditional deep learning using point estimation. In contrast, the generator that introduces Bayesian learning observes a more comprehensive distribution sampling of the parameters. This distribution sampling can help us choose better generator parameters for image fusion. To illustrate, compared to the other two results, the second image fusion result in Figure. 1 reduces the image noise in the marked area.

\subsection{Network design}
\begin{figure}  
	\centering  
	\includegraphics[width=15cm]  {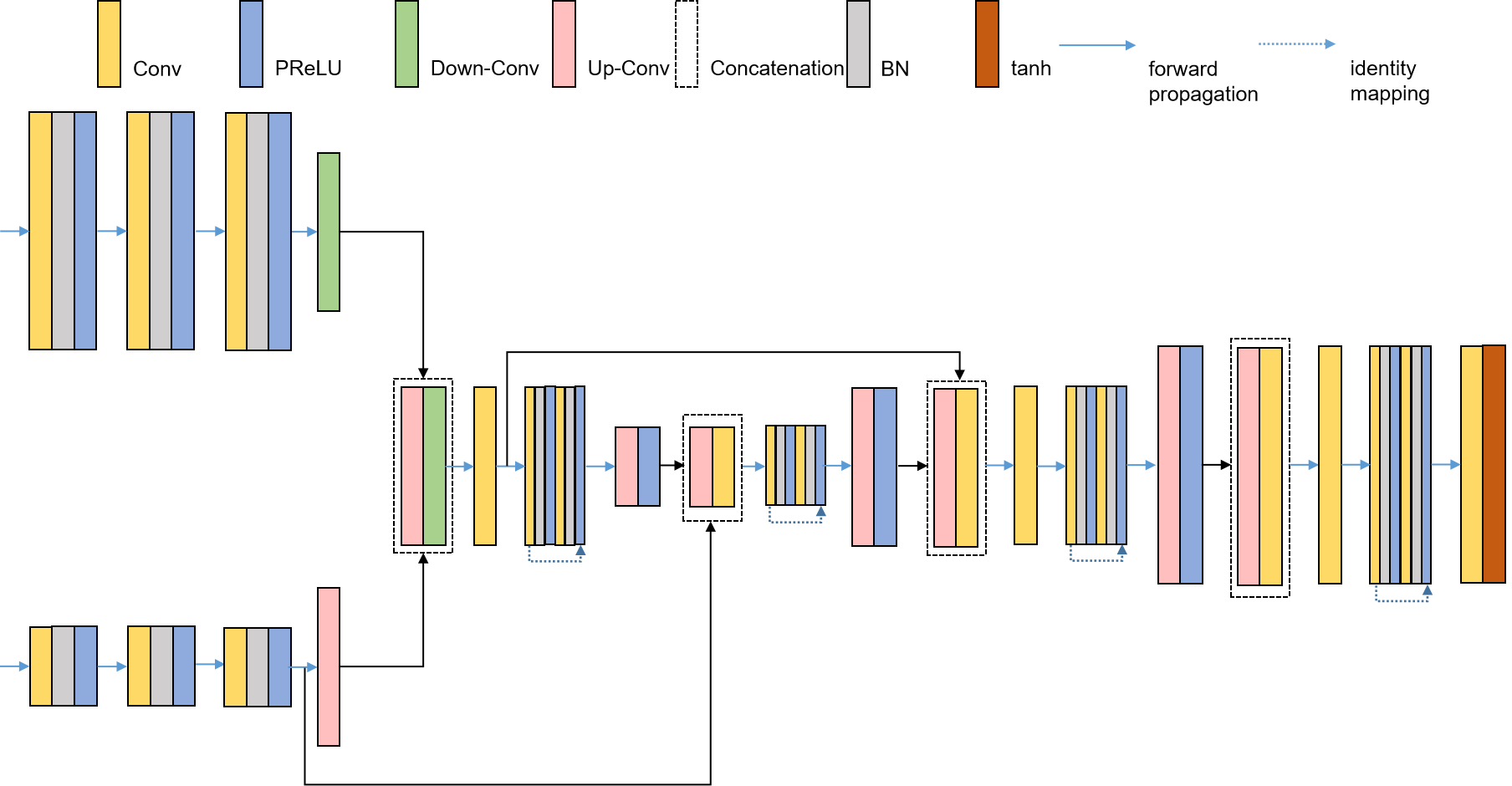} 
	\caption{\label{2}  Architecture of proposed generator network}
	\label{Fig2}
\end{figure}

In this subsection, we describe the architectural design of the PSGLD-BGAN. Figure. \ref{Fig2} describes in detail the design of the two-stream generative network of the PSGLD-BGAN, including feature extraction, feature fusion, and image reconstruction. For the discriminator network, we use a Markovian discriminator \citep{p27}, which is commonly used in the field of image generation. When using the worldview dataset as an input, Table \ref{tab1} and Table \ref{tab2} show the network design parameters for the generative and adversary networks.

\begin{table*}[!htbp]
	\caption{\label{tab1} Parameters of proposed generator network}
	\begin{tabular}{@{}ccccc@{}}
		\toprule
		Module               &           & Layer       & Kernel/Stride & Output                          \\ \midrule
		Input                & PAN       &             &                              & $400\times 400\times 4$         \\
		& MS        &             &                              & $100\times 100\times 4$         \\
		& Reference &             &                              & $400\times 400\times 4$         \\
		Feature Extraction   & PAN       & Conv1\_P    & $3\times 3\times 32/1$       & $400\times 400\times 32$        \\
		&           & Conv2\_P    & $3\times 3\times 32/1$       & $400\times 400\times 32$        \\
		&           & Conv3\_P    & $3\times 3\times 32/1$       & $400\times 400\times 32$        \\
		&           & Down\_Conv1 & $2\times 2\times 64/2$       & $200\times 200\times 64$        \\
		& MS        & Conv1\_M    & $3\times 3\times 32/1$       & $100\times 100\times 32$        \\
		&           & Conv2\_M    & $3\times 3\times 32/1$       & $100\times 100\times 32$        \\
		&           & Conv3\_M    & $3\times 3\times 32/1$       & $100\times 100\times 32$        \\
		&           & Up\_Conv1   & $2\times 2\times 64/2$       & $200\times 200\times 64$        \\
		Feature Fusion       &           & Concat1     &                              & $200\times 200\times (64+64)$   \\
		&           & Conv4       & $3\times 3\times 128/1$      & $200\times 200\times 128$       \\
		&           & Conv5       & $3\times 3\times 128/1$      &                                 \\
		&           & Conv6       & $3\times 3\times 128/1$      &                                 \\
		&           & Down\_Conv2 & $2\times 2\times 256/2$      & $100\times 100\times 256$       \\
		&           & Concat2     &                              & $100\times 100\times (256+32)$  \\
		&           & Conv7       & $1\times 1\times 256/1$      & $100\times 100\times 256$       \\
		&           & Conv8       & $3\times 3\times 256/1$      & $100\times 100\times 256$       \\
		&           & Conv9       & $3\times 3\times 256/1$      & $100\times 100\times 256$       \\
		Imgae Reconstruction &           & Up\_Conv2   & $2\times 2\times 128/2$      & $200\times 200\times 128$       \\
		&           & Concat3     &                              & $200\times 200\times (128+128)$ \\
		&           & Conv10      & $1\times 1\times 256/1$      & $200\times 200\times 256$       \\
		&           & Conv11      & $3\times 3\times 256/1$      & $200\times 200\times 256$       \\
		&           & Conv12      & $3\times 3\times 256/1$      & $200\times 200\times 256$       \\
		&           & Up\_Conv3   & $2\times 2\times 128/2$      & $400\times 400\times 128$       \\
		&           & Concat4     &                              & $400\times 400\times (128+32)$  \\
		&           & Conv13      & $1\times 1\times 64/1$       & $400\times 400\times 64$        \\
		&           & Conv14      & $3\times 3\times 64/1$       & $400\times 400\times 64$        \\
		&           & Conv15      & $3\times 3\times 64/1$       & $400\times 400\times 64$        \\
		&           & Conv16      & $3\times 3\times 4/1$        & $400\times 400\times 4$         \\
		&           &             &                              & $400\times 400\times 4$         \\ \bottomrule
	\end{tabular}
\end{table*}

\begin{table*}[!htbp]
\caption{\label{tab2}  Parameters of proposed generator network}
	\begin{tabular}{ccccc}

\toprule
Module &                             & layer   & Kernel/Stride          & Output                   \\ \midrule
Input  & Reference image (True data) &         &                        & $400\times 400\times 4$  \\
& Fusion results (False data) &         &                        & $400\times 400\times 4$  \\
&                             & Conv1   & $3\times 3\times 64/2$ & $200\times 200\times 64$ \\
&                             & Conv2   & $3\times 3\times 16/2$ & $100\times 100\times 16$ \\
&                             & Conv3   & $3\times 3\times 4/2$  & $50\times 50\times 4$    \\
&                             & Conv4   & $3\times 3\times 4/2$  & $25\times 25\times 4$    \\
&                             & Conv5   & $3\times 3\times 1/1$  & $25\times 25\times 1$    \\
&                             & Sigmoid &                        & $25\times 25\times 1$    \\ \bottomrule
	\end{tabular}
\end{table*}
\subsection{Loss function} 
In addition to the network architecture, the loss function is another important factor that affects the quality of the reconstructed image. Several works \citep{p28,p29} have shown that $l_11$ loss is a better choice when performing image recovery tasks. Hence in this work, we also leverage $l_11$ as part of the generator loss to train our network. Given a set of training samples $\left\{ X_P^i,X_M^i,Y^i  \right\} $, where $X_P^i$ and $X_M^i$ are the PAN image and low resolution MS image, respectively, and $Y^i$ is the corresponding reference image. Then, $l_1$ loss is defined as:
\begin{equation}
\label{e10}
l_1=\frac{1}{N}\sum_{i=1}^N{\left| F\left( X_{P}^{i},X_{M}^{i} \right) -Y^i \right|_1}
\end{equation}
Thus, the loss of PSGLD-BGAN generator and discriminator networks can be respectively described as follows:
\begin{equation}
\label{e11}
l_g=E_{z\sim p_z\left( z \right)}\left[ \log \left( 1-D\left( G\left( z \right) \right) \right) \right] +l_1
\end{equation}
\begin{equation}
\label{e12}
l_d=-E_{x\sim p_{data}\left( x \right)}\left[ \log D\left( x \right) \right] -E_{z\sim p_z\left( z \right)}\left[ \log \left( 1-D\left( G\left( z \right) \right) \right) \right]
\end{equation}
where N is the number of training samples in a mini-batch and $f$ represents the proposed image fusion model.

\section{Experiments and analysis}
\subsection{Data sets}
We experiment on data sets from QuickBird and WorldView\footnote{http://www.digitalglobe.com/.} to validate the effectiveness of the proposed PSGLD-BGAN on the image-fusion (Pan-Sharping) mission. Table \ref{tab3} details the spatial resolution and wavelength range of the two remote sensing satellites. The data in this paper are part of the Roman city images of Italy taken by QuickBird and part of the Mexican city images taken by WorldView.
The scale ratio of low spatial resolution MS image to high spatial resolution PAN image is 4:1. Therefore, Therefore, we downsampled the original single-band PAN and quad-band (Red, Green, Blue and Near infrared) MS images with one-quarter contraction and used the original MS images as reference images. To accommodate the training and testing of the network, for Worldview satellite data, we set the input PAN image (QuickBird data) size to $400\times 400\times 4$, the MS image size to $100\times 100\times 4$, and the reference image size $400\times 400\times 4$; for QucikBird satellite data input PAN image (WorldView data) size is $512\times 512\times 4$, MS image size is $128\times 4 \times 128 \times 4$ and reference image size $512 \times 512 \times 4$. 

\begin{table*}[!htbp]
	\caption{\label{tab3} Spectral and spatial characteristics of PAN and MS images from Qucikbird and WorldView}
	\begin{center}	
	\begin{tabular}{lcccccc}
		\hline
		\multirow{2}{*}{Satellite}     & \multicolumn{4}{c}{Spectral wavelength(nm)} & \multicolumn{2}{c}{Spatial resolution(m)} \\ \cline{2-7} 
		& Red       & Green     & Blue     & Nir      & PAN                 & MS                  \\ \hline
		\multicolumn{1}{c}{Quickbird}  & 630-690   & 520-600   & 450-520  & 760-890  & 0.6                 & 2.4                 \\ \hline
		\multicolumn{1}{c}{WordView-4} & 655-690   & 510-580   & 450-510  & 780-920  & 0.31                & 1.24                \\ \hline
	\end{tabular}
 	\end{center}
\end{table*}
\subsection{Evaluation indexes}
We employ five popular indexes to evaluate the performance of several classic methods and advanced ones. 

\textbf{CC} The Correlation Coeffient (CC) is the similarity between fusion image and reference image, which is mainly used to observe the spectral quality of fusion images through pan-sharping. The value range of CC is [0,1], and the spectral quality comes to be better with the value increasing.
\begin{equation}
\label{e13}
CC=\frac{\sum_{i=1}^W{\sum_{j=1}^H{\left( X_{i,j}-m_X \right) \left( Y_{i,j}-m_Y \right)}}}{\sqrt{\sum_{i=1}^W{\sum_{j=1}^H{\left( X_{i,j}-m_X \right) ^2\sum_{i=1}^W{\sum_{j=1}^H{\left( Y_{i,j}-m_Y \right) ^2}}}}}}
\end{equation}
where X and Y represent the fusion image and the reference image respectively, m is the mean value of an image.

\textbf{UIQI} \citep{p30}The universal image quality index (UIQI) is an index capturing the degree of structural distortion of the fusion image, which can be defined as:
\begin{equation}
\label{e14}
UIQI=\frac{\sigma _{XY}}{\sigma _X\sigma _Y}\times \frac{2\mu _X\mu _Y}{\mu _X\mu _Y}\times \frac{2\sigma _X\sigma _Y}{\sigma _X+\sigma _Y}
\end{equation}
where $\sigma_ {XY}$ and $\mu_{XY}$ represent standard variance and mean of image X and image Y and the similar definition can be employed in $ \sigma_ {X}$ ,$ \sigma_ {Y}$, $ \mu_{X}$ and $ \mu_ {Y}$.   

\textbf{SAM} \citep{p31} The spectral angle mapper(SAM) can measure spectral distortions between fusion results through pan-sharpening and the corresponding reference images. It is defined as:
\begin{equation}
\label{e15}
SAM\left( x_P,x_R \right) =\arccos \left( \frac{x_P\cdot x_R}{\lVert x_P \rVert \cdot \lVert x_R \rVert} \right) 
\end{equation}
where $x_P$ and $x_R$ respectively represent the spectral vectors from fusion images and reference images.

\textbf{ERGAS} \citep{p32} The  erreur relative globale adimensionnelle de synth\`ese (ERGAS) measures the global dimensional synthesis error, which can be calculated as:
\begin{equation}
\label{e16}
ERGAS=100\frac{h}{l}\sqrt{\frac{l}{K}\sum_{k=1}^K{\left( \frac{rmse_i}{m_i} \right)}}
\end{equation}
where $h$ and $l$ represent the spatial resolution of PAN and MS images, respectively; $rmse_i$ is the root mean square error between the $i$th band of the fusion image and reference image; $m_i$ is the mean value of the $i$th band of the MS image.

$\boldsymbol{Q}_4$ \citep{p33} The quality-index $Q_4$ is the 4-band extension of Q index. $Q_4$ can be calculated as:
\begin{equation}
\label{e20}
Q_4=\frac{4\left| \sigma _{z_1z_2} \right|\cdot \left| \mu _{z_1} \right|\cdot \left| \mu _{z_2} \right|}{\left( \sigma _{z_1}^{2}+\sigma _{z_2}^{2} \right) \left( \mu _{z_1}^{2}+\mu _{z_2}^{2} \right)}
\end{equation}
where $z_1$ and $z_2$ are two quaternions, which consist of spectral vectors of MS images, $\mu _{z_21}$ and $\mu _{z_2}$ are the mean value of $z_1$ and $z_2$, denotes the variances of z1 and z2, and $\sigma _{z_1z_2}$ represents the covariance between$z_1$ and $z_2$.

\subsection{Implement environment}
Through data augmentation, we obtain two final data sets, i.e., the QuickBird data set with 12,000 pairs (PAN, MS and reference image) and the WorldView data set with 10,000 pairs, respectively. To avoid random results, we randomly utilize 80\% of the data for training and the remainder for testing. The proposed model is implemented in PyTorch and run on a GTX 1080TI. The loss is minimized through the proposed optimizer based on PSGLD strategy with a learning rate 0.0001. The mini-batch is set to 16. During the training process, we acquire 50 posterior samples of the generator parameter. And the generator parameter with the highest CC index is taken as the final selected parameter. 

\subsection{Comparison with other models}
In this subsection, we compare the proposed method with state of the art deep learning models such as MSDCNN \citep{p36}, PanNet \citep{p16} ,ResTFNet \citep{p14} and Pan-GAN \citep{p37}.

\begin{figure} [!htbp] 
	\centering  
	\includegraphics[width=10cm]  {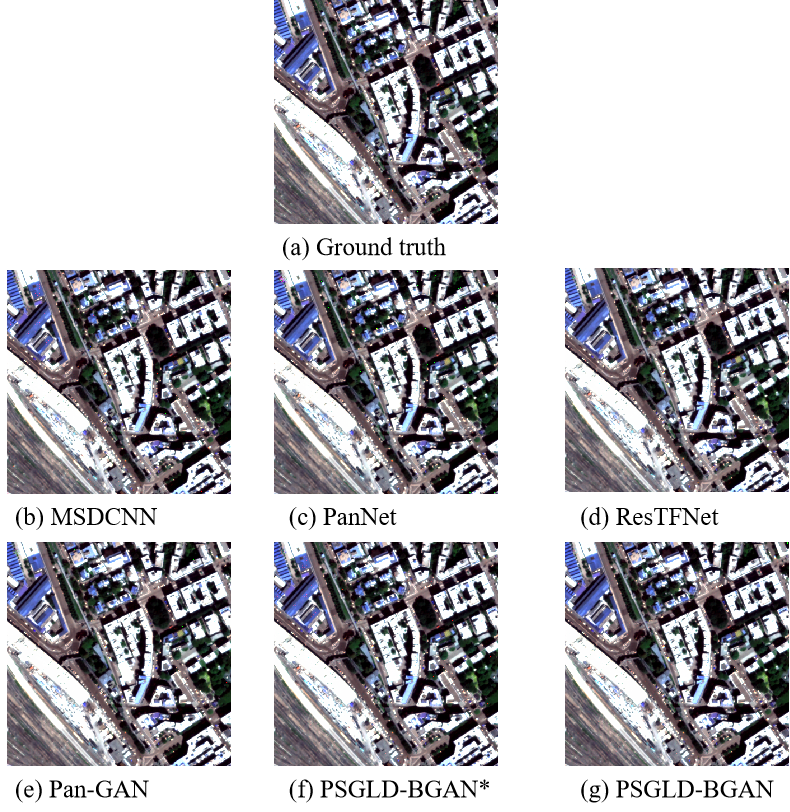} 
	\caption{\label{3} Pansharpening results on the Quickbird images}
	\label{Fig3}
\end{figure}

\begin{figure}  
	\centering  
	\includegraphics[width=10cm]  {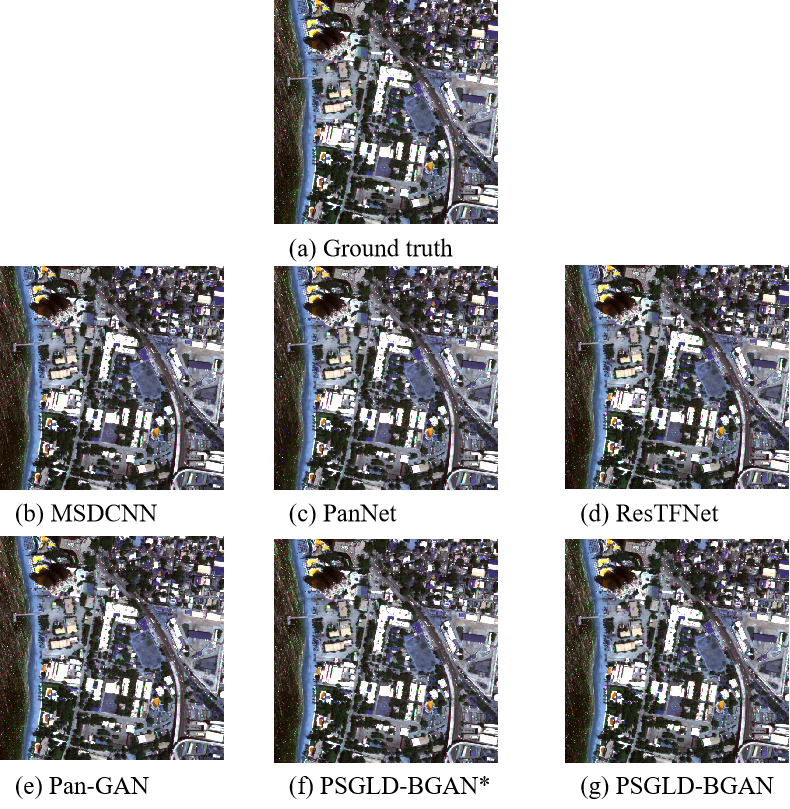} 
	\caption{\label{4} Pansharpening results on the WordView images}
	\label{Fig4}
\end{figure}

Table \ref{tab4} and Table \ref{tab5} respectively show the quantitative evaluation of experiments on Quickbird and WorldView data sets. Figure \ref{Fig3} and Figure \ref{Fig4}  illustrate examples of comparison methods from two data sets. PSGLD-BGAN* represents the proposed model without bayesian learning, sharing the same architecture and parameter. 

From Tables \ref{tab4} and \ref{tab5}, we can observe that the traditional deep learning version, PSGLD-BGAN*, is already capable of rivaling recent advanced models. Further, PSGLD-BGAN incorporates Bayesian learning for posteriori sampling, selects the final parameter in a simple way, and achieves the best performance on all evaluation metrics. 

\begin{table*}[!htbp]
	\caption{\label{tab4} Quantitative assessment on the Quickbird data sets}
	\begin{center}	
	\begin{tabular}{cccccc}
		\toprule
		& CC$\uparrow $ & UIQI$\uparrow$ & SAM$\downarrow$ &  ERGAS$\downarrow$ & Q4$\uparrow$ \\\midrule
		MSDCNN                    & 0.9352        & 0.9483         & 4.2387          & 5.5578            & 0.9403          \\
		PanNet                    & 0.9472        & 0.9523         & 4.1345          & 5.2014            & 0.9499          \\
		ResTFNet                  & 0.9828        & 0.9843         & 3.8775          & 4.1544            & 0.9822          \\
		Pan-GAN                   & 0.9832        & 0.9827         & 3.9564          & 4.0313            & 0.9831          \\
		PGSLD-BGAN*               & 0.9821        & 0.9833         & 3.8211          & 4.0154            & 0.9817          \\
		PSGLD-BGAN                & 0.9899        & 0.9853         & 3.7078          & 3.7154            & 0.9895         \\\bottomrule
	\end{tabular}
	\end{center}
\end{table*}

\begin{table*}[!htbp]
	\caption{\label{tab5} Quantitative assessment on the WorldView data sets}
	\begin{center}	
	\begin{tabular}{cccccc}
		\toprule
		& CC$\uparrow $ & UIQI$\uparrow$ & SAM$\downarrow$ & ERGAS$\downarrow$ & Q4$\uparrow$ \\\midrule
		MSDCNN                    & 0.9423        & 0.9311         & 4.4984          & 4.2457            & 0.9256          \\
		PanNet                    & 0.9478        & 0.9422         & 4.0345          & 4.1476            & 0.9405          \\
		ResTFNet                  & 0.9802        & 0.9821         & 3.6664          & 4.0954            & 0.9834          \\
		Pan-GAN                   & 0.9811        & 0.9822         & 3.6743          & 3.8922            & 0.9846          \\
		PGSLD-BGAN*               & 0.9825        & 0.9817         & 3.7089          & 3.9841            & 0.9820          \\
		PSGLD-BGAN                & 0.9887        & 0.9836         & 3.6103          & 3.7548            & 0.9875         \\\bottomrule
	\end{tabular}
	\end{center}
\end{table*}

\section{Conclusion}
In this paper, we propose a Bayesian Generative Adversarial Network based on Preconditioned Stochastic Gradient Langevin Dynamics for solving remote sensing pan-sharpening. Deep learning techniques, especially the GAN architecture, which severe as a powerful and flexible tool, have gradually became the dominant framework in the field of image generation. We design a generator network to extract the features of PAN and MS images and fuse them together naturally for pan-sharpening. Then, we distinguish the fusion images from the reference images through proposed discriminator network to enhance the image fusion performance. Different from traditional image fusion deep learning models, we use PSGLD strategy to sample the posterior parameter of the generator network, expand the parameter space to explore the generator network, and select the best generator parameter among them. In the future, we will develop a Bayesian deep generative framework that is more suitable for image fusion of people.

\bibliographystyle{unsrt}

\bibliography{PSGLDBGAN.bib}

\end{document}